\documentclass[conference]{IEEEtran}
\IEEEoverridecommandlockouts
\newcommand{\balg}{\begin{algorithm}}
\newcommand{\ealg}{\end{algorithm}}
\usepackage{algorithmic}
\usepackage[ruled,vlined,linesnumbered]{algorithm2e}

\def\BibTeX{{\rm B\kern-.05em{\sc i\kern-.025em b}\kern-.08em
    T\kern-.1667em\lower.7ex\hbox{E}\kern-.125emX}}
\usepackage{microtype}
\usepackage[T1]{fontenc}
\usepackage[utf8]{inputenc}
\usepackage{amssymb}
\usepackage[english]{babel}
\usepackage{float}
\usepackage{enumitem}
\usepackage{tabularx}
\usepackage{booktabs}

\usepackage[sorting=none,firstinits=true,maxbibnames=3]{biblatex}
\usepackage{csquotes}
\usepackage[table]{xcolor}%
\usepackage{commath}
\usepackage{url}
\usepackage[colorlinks=false]{hyperref}
\usepackage{amsmath,amssymb,amsfonts}   

\bibliography{main.bib}
\usepackage[normalem]{ulem}
\usepackage{comment}
\usepackage[nameinlink,noabbrev]{cleveref}
\usepackage{mathtools}
\usepackage{placeins}
\usepackage{glossaries}

\usepackage[T1]{fontenc}
\usepackage{lmodern}

\usepackage{caption}
\usepackage{subcaption}

\usepackage{paralist}

\newacronym{PM}{PM}{Process Mining}
\newacronym{BPM}{BPM}{Business Process Management}
\newacronym{ML}{ML}{Machine Learning}
\newacronym{DM}{DM}{Data Mining}
\newacronym{LHS}{LHS}{Left-Hand-Side}
\newacronym{RHS}{RHS}{Right-Hand-Side}
\newacronym{CNN}{CNN}{Convolutional Neural Network}
\newacronym{DK}{DK}{Deterministically known}
\newacronym{IM}{IM}{Inductive Miner}
\newacronym{SK}{\emph{SK}}{stochastically known}
\newacronym{S-ABCC}{\emph{S-ABCC}}{Stochastic Alignment-Based Conformance Checking}
\newacronym{SKTR}{\emph{SKTR}}{Stochastically Known Trace Recovery}
\newcommand{\algNameNew}{\emph{SKTR}}
\newcommand{\algNameNewSpace}{\emph{SKTR }} 

\begin{document}

\title{SKTR: Trace Recovery from Stochastically Known Logs
% \thanks{}
}
% \author{\IEEEauthorblockN{Eli Bogdanov}
% \IEEEauthorblockA{\textit{Industrial Engineering \& Management} \\
% \textit{Technion--Israel Institute of Technology}\\
% Haifa 3200003, Israel \\
% eli-bogdanov@campus.technion.ac.il\\
% orcidID: 0000-0002-5084-7344}
% \and
% \IEEEauthorblockN{Izack Cohen}
% \IEEEauthorblockA{\textit{Faculty of Engineering} \\
% \textit{Bar-Ilan University}\\
% Ramat Gan 5290002, Israel \\
% izack.cohen@biu.ac.il\\
% orcidID: 0000-0002-6775-3256}
% \and
% \IEEEauthorblockN{Avigdor Gal}
% \IEEEauthorblockA{\textit{Industrial Engineering \& Management} \\
% \textit{Technion--Israel Institute of Technology}\\
% Haifa 3200003, Israel \\
% avigal@technion.ac.il\\
% orcidID: 0000-0002-7028-661X}
% }

\author{\IEEEauthorblockN{Eli Bogdanov}
\IEEEauthorblockA{\textit{Data \& Decision Sciences} \\
\textit{Technion--Israel Institute of Technology}\\
Haifa 3200003, Israel \\
eli-bogdanov@campus.technion.ac.il\\
orcidID: 0000-0002-5084-7344}
\and
\IEEEauthorblockN{Izack Cohen}
\IEEEauthorblockA{\textit{Faculty of Engineering} \\
\textit{Bar-Ilan University}\\
Ramat Gan 5290002, Israel \\
izack.cohen@biu.ac.il\\
orcidID: 0000-0002-6775-3256}
\and
\IEEEauthorblockN{Avigdor Gal}
\IEEEauthorblockA{\textit{Data \& Decision Sciences} \\
\textit{Technion--Israel Institute of Technology}\\
Haifa 3200003, Israel \\
avigal@technion.ac.il\\
orcidID: 0000-0002-7028-661X}
}

\maketitle

\begin{abstract}
Developments in machine learning together with the increasing usage of sensor data challenge the reliance on deterministic logs, requiring new process mining solutions for uncertain, and in particular stochastically known, logs. In this work we formulate {\em trace recovery}, the task of generating a deterministic log from stochastically known logs that is as faithful to reality as possible. An effective trace recovery algorithm would be a powerful aid for maintaining credible process mining tools for uncertain settings. We propose an algorithmic framework for this task that recovers the best alignment between a stochastically known log and a process model, with three innovative features. Our algorithm, $SKTR$,  1) handles both Markovian and non-Markovian processes; 2) offers a quality-based balance between a process model and a log, depending on the available process information, sensor quality, and machine learning predictiveness power; and 3) offers a novel use of a synchronous product multigraph to create the log. 
An empirical analysis using five publicly available datasets, three of which use predictive models over standard video capturing benchmarks, shows an average relative accuracy improvement of more than $10\%$ over a common baseline.

\end{abstract}

%\begin{IEEEkeywords}
%component, formatting, style, styling, insert
%\end{IEEEkeywords}

\section{Introduction} \label{sec:intro}
Process logs constitute an essential element of process mining tasks.~\footnote{\url{https://www.processmining.org/event-data.html\#data}} 
They serve as the basis to process discovery, enable conformance checking, and their reliability determines to a great extent the success of process enhancement. Process log capturing has seen a great transformation in recent years, from a meticulous update in an information system using the caring hands of human clerks to semi and fully automatic activity capturing from sensors, keyboard strokes, and video clips~\cite{9234741}. Such capturing methods challenge the common reliance of process mining on deterministic logs~\cite{cohen2021uncertain}. For example, consider log discovery from video clips. Activity identification in video clips using deep learning is typically based on choosing an activity with the highest likelihood from a prediction layer called the softmax layer, which provides a distribution over possible activities. By deterministically selecting a single activity, even if its likelihood is higher than other possible activities, the log becomes a biased representation of reality.

%, highlights the need to explore non-deterministic logs and motivates the present research. %Given a video clip, a classification task can identify recorded process activities ({\em e.g.}, making coffee or preparing a salad). 
%Such classification  
A recent line of work considers uncertain event logs~\cite{pegoraro2019discovering, cohen2021uncertain, bogdanov2022conformance}, where uncertainty can be associated with timestamps, activities, {\em etc}. Going back to the video capturing example, using \gls{SK} logs~\cite{cohen2021uncertain}, the softmax layer can serve as input, where activities in a trace are assigned a probability value, approximated using the softmax layer.

Modeling process logs as \gls{SK} logs leads to the challenge we tackle in this work, that of identifying a true trace out of several possible realizations, each of which can be associated with different managerial and operational implications ({\em e.g.}, resource allocation and decision making). Therefore, credible decision making vis-\`a-vis process management requires recovering the original event-log traces. 

In this work we focus on {\em trace recovery}, a challenging process mining task~\cite{beerepoot2023biggest} involving the accurate selection of the log realization that represents the true occurrence from an \gls{SK} log.  
%We do so by  using the context of an activity within a process. 
Accordingly, given a process model and an \gls{SK} log, we introduce the \gls{SKTR} algorithm that recovers deterministic traces with the aim of maximizing the similarity between a recovered log and a log that represents the actual occurrence of activities in reality. A challenging aspect of this approach involves the quality of the process models, both conceptual and data-driven~\cite{10.1007/978-3-319-45348-4_11}. Process models depend heavily on the inputs and parameters that shape them, requiring modelers' domain expertise and high quality of execution data. Designer's choice of a discovery algorithm also affect the model's quality. To overcome this hurdle, we introduce a flexible approach that can align a trace recovery process with the amount of trust we assign to the available reference process model and the \gls{SK} log.

This work makes a threefold contribution:
\begin{compactenum}[(1)]
	    \item From a modeling viewpoint (Section~\ref{sec:model}), the suggested approach can represent varying levels of trust in the process model and the log. We show that in stochastic settings with imprecise data, the role of process models is more important for recovery of the original log. Our approach uses cost functions to reflect levels of trust in the model and log, considering frequent and infrequent model paths. %in the context of conformance checking over \gls{SK} traces. 
    We discuss how to map probabilities into costs and explore the properties and performance of several linear and non-linear cost functions.  
    \item From an algorithmic perspective (Section~\ref{sec:SABCCAndTraceRecovery}), we develop \gls{SKTR}, a novel approach to recover an \gls{SK} log, using a probability-aware conformance checking algorithm~\cite{bogdanov2022conformance}. We propose a novel use of a synchronous product multigraph to capture alternative trace, model, and synchronous moves as well as their probabilities. \gls{SKTR} handles Markovian and non-Markovian (path dependent) probabilities, where the latter can improve the recovery accuracy.  Through analytical insights and an empirical evaluation, we demonstrate that \gls{SKTR} achieves high accuracy in trace recovery under varying conditions and that it is highly preferable compared to several machine learning and process mining baselines.
    \item From an empirical viewpoint (Section~\ref{sec:experimental_study}), we demonstrate an average relative accuracy improvement of more than $10\%$ over a common baseline and analyze the impact of various parameters, such as discovered process model quality and the stochastic noise within the event log, on recovery accuracy. 
\end{compactenum}
Related work is discussed in Section~\ref{sec:related_work} and Section~\ref{sec:conclusions} concludes the paper and discusses future research directions.

\section{Model}\label{sec:model}
We start by describing an \gls{SK} environment and position our search for a cost model within this context. We adopt the notation of \citeauthor{cohen2021uncertain} \cite{cohen2021uncertain} who characterized environments of interest as either \gls{DK} or \gls{SK}. \gls{DK} refers to a process model or an event log that are deterministic ({\em e.g.}, a supervised dataset), and \gls{SK} to a known probability distribution of event attribute values in an event log ({\em e.g.}, predicted activity names for a trace event). Accordingly, for an \gls{SK} trace, the probability distribution of each event to be classified as one of the possible activities is known. 

\begin{table}[htpb]
	\centering
%         \begin{adjustbox}{width=0.5\textwidth}
	\begin{tabular}{c|c|c|c}
		\textbf{Case\#} & \textbf{Event\#} & \textbf{Activity} & \textbf{Timestamp} \\ 
		\hline
		1 & $e_1$ & $\{A:0.8, B:0.2\}$ & 03-06-2022T12:00  \\ 
		1 & $e_2$ & $\{C:0.7, D:0.3\}$ & 03-06-2022T14:55  \\ 
		1 & $e_3$ & $\{E:0.6, F:0.4\}$ & 04-06-2022T17:39
		\vspace{5pt}
	\end{tabular}
%\end{adjustbox}
\caption{\label{tab:stochastic_trace_for_the_running_example} An excerpt from an \gls{SK} log}
		\vskip-.3in
\end{table}

\begin{figure}[htpb]
	\begin{center}	
		\includegraphics[ width=0.45\textwidth, height=1in]{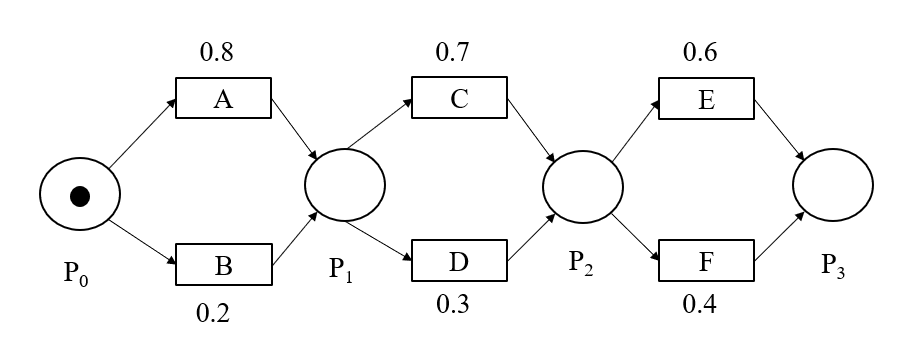}
		\caption{The stochastic log model of the \gls{SK} log in Table~\ref{tab:stochastic_trace_for_the_running_example}}
		\label{fig:stochastic log model example}
	\end{center}
		\vskip-.15in
\end{figure}

Table~\ref{tab:stochastic_trace_for_the_running_example} presents an excerpt from an \gls{SK} log. Each entry consists of a case identifier, an event identifier, a probabilistic distribution over possible activities, and a timestamp. For example, event $e_1$ may relate to activity $A$ with a probability of $0.8$ and activity $B$ with a probability of $0.2$. The corresponding stochastic trace model (see~\cite{bogdanov2022conformance}, Definition~1) is presented in Figure~\ref{fig:stochastic log model example}. 

In this work, we seek to recover traces using a process model that can be either human generated or discovered from an event log. For the latter, we shall assume that the model was discovered from a \gls{DK} log, where no uncertainty exists. \citeauthor{10.1007/978-3-319-45348-4_11}~\cite{10.1007/978-3-319-45348-4_11} highlighted the role of process quality in conformance checking and, in line with this reasoning, we argue that in stochastic settings, recovered traces depend heavily on the quality of the process model, with a trade-off between fitness and precision. A low fitting model can result in rejecting true activity labels within a non conforming trace. A low precision model can result in recovering wrong activity labels if such activities are possible according to the model. 
The recovered trace depends also on data quality. On one extreme, if the quality of sensor data is high than data can be trusted and the importance of a reference process model decreases. On the other extreme, if data are imprecise and the model is of high quality, the role of the model in determining the true trace increases. Oftentimes, we place our trust somewhere between these extremes.

Given our aim to recover deterministic traces from an \gls{SK} log and our intention to use possibly unreliable process models and data, we seek a mechanism to control the reliance of trace recovery algorithms on process models, based on their quality. Consider stochastic conformance checking algorithms~\cite{bogdanov2022conformance}, where for each set of alternative transitions that represent nonsynchronous trace moves, the algorithm moves forward with a single {\bf randomly chosen} transition while ignoring the remaining transitions. The random selection among alternative transitions stems from the fact that the default weight of nonsynchronous moves in conformance checking is set to 1 (both in deterministic and stochastic settings, see~\cite{bogdanov2022conformance}). Therefore, no transition that forms nonsynchronous moves is preferred, regardless of its probability to be part of a more likely activity sequence. Ignoring alternative nonsynchronous moves, however, may influence the trace recovery accuracy. 

\begin{figure}[htpb]
	\centering
        \begin{subfigure}[b]{0.48\textwidth}
        \centering
		  \includegraphics[width=\textwidth]{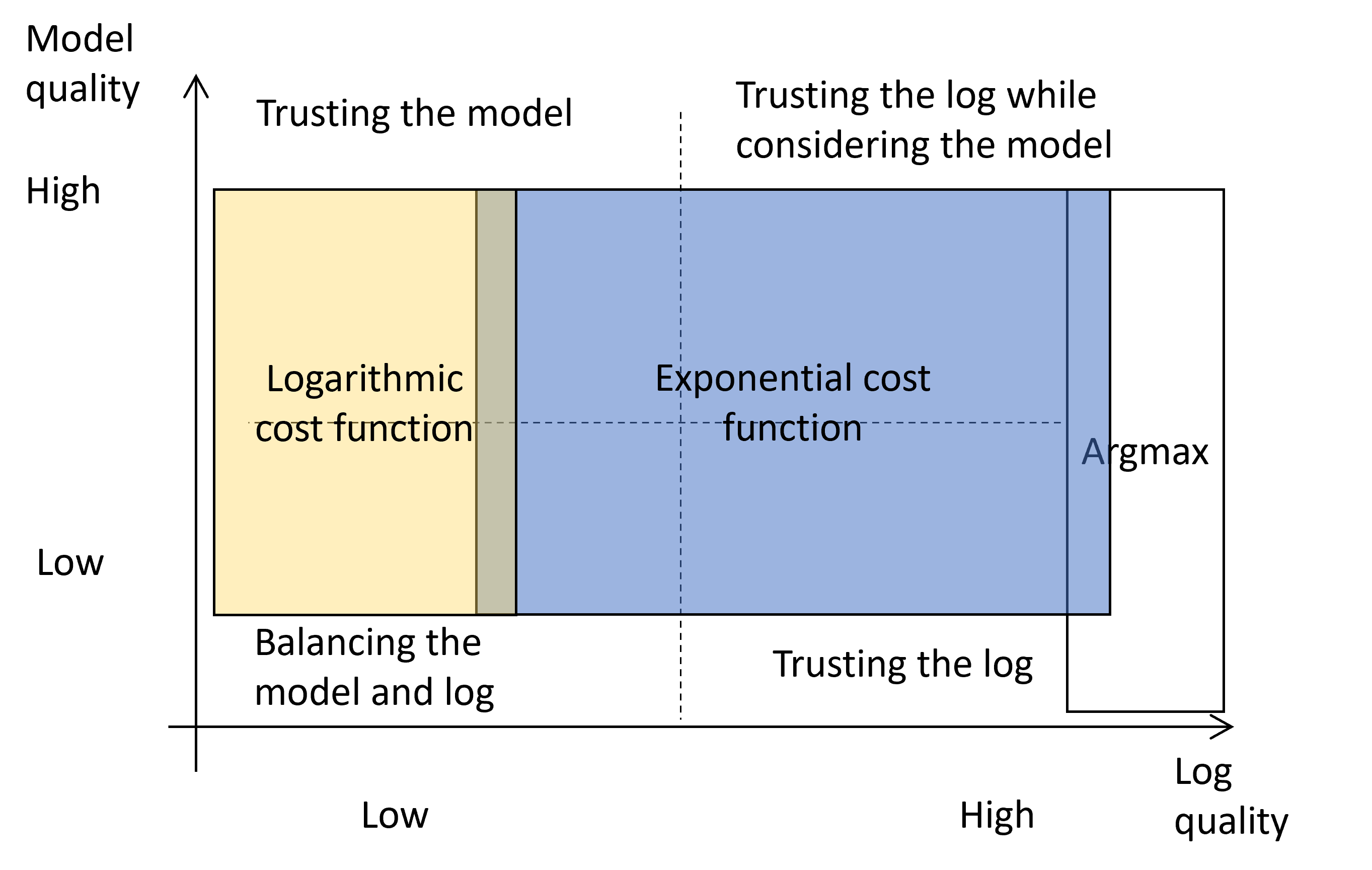}
		  \caption{}
		  \label{fig:model_vs_log_trust}
        \end{subfigure} 
	\hfill   
        \begin{subfigure}[b]{0.48\textwidth}
        \centering
		  \includegraphics[width=\textwidth]{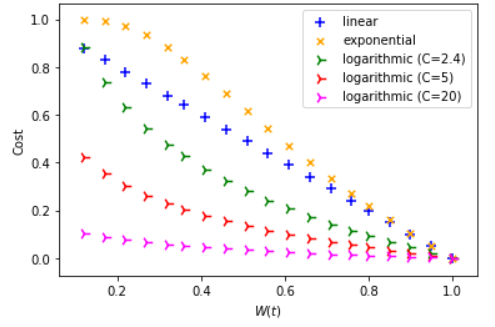}
		  \caption{}
		  \label{fig:cost_func_trustworth}
            \end{subfigure} 
            \caption{(a) Types of cost functions and their relation to reliance on the model vs. the log; (b) Cost as a function of the probability for some functions.}
		\vskip-.25in
\end{figure}

We use cost functions to reflect the trust in the model {\em vs.} the stochastic trace (Figure~\ref{fig:model_vs_log_trust}) and calculate the cost value by each function as a function of the probability of an activity (Figure~\ref{fig:cost_func_trustworth}). Today, Argmax is the most commonly used heuristic for generating a prediction from a softmax layer. Our cost functions, however, eliminate the need to use an Argmax heuristic that ignores the process model by choosing, for each event in an \gls{SK} trace, the activity label associated with the highest probability. Had we applied an Argmax heuristic on the \gls{SK} trace in Table \ref{tab:stochastic_trace_for_the_running_example} we would have generated the trace $\langle A,C,E\rangle$. 

%\begin{sloppypar}
\citeauthor{bogdanov2022conformance}~\cite{bogdanov2022conformance} suggested an exponential cost function $Weight(e) = 1 - e^{1- \frac{1}{W(t)}}$ for stochastic conformance checking, where $Weight(e)$ and $W(t)$ are the cost of an edge on the reachability graph and the probability of a transition $t$, respectively. 
In addition, they introduced a linear cost function $Weight(e) = 1 - W(t)$, and a logarithmic function $Weight(e)= \frac{-ln(W(t))}{C}$, where $C$ is a normalization constant such that costs that correspond to a synchronous move do not exceed $1$. The difference between these functions, as illustrated in Figure~\ref{fig:cost_func_trustworth}, is the rate at which they induce an additional cost as the probability of a transition decreases. The costs induced by the exponential and linear functions are similar for higher probabilities ({\em e.g.}, $W(t) \geq 0.8$) 
while for smaller probabilities ({\em e.g.},  $0.2 \leq W(t) \leq 0.8$) the exponential function penalizes uncertainty to a larger degree. As the probability approaches $0$, the penalty given by both functions approaches $1$. In contrast, the normalized logarithmic function is more ``forgiving,'' %(see Figure~\ref{fig:model_vs_log_trust}), 
penalizing uncertainty to a lesser degree (on average) compared to the other two functions, thus trusting the model more. The constant factor $C$ can be used to fine-tune the ``forgiveness'' of the cost function since the penalty magnitude decreases with $C$ values across all probability values. %We note that using new cost functions or changing the structure of the above-mentioned cost functions may change their behaviour and the \gls{SKTR} performance, but this is beyond the focus of this paper.
%\end{sloppypar}
\section{SKTR Algorithm} \label{sec:SABCCAndTraceRecovery}
We now introduce \gls{SKTR} (Algorithm~\ref{alg:sktr}), %our proposed algorithmic framework. Algorithm~\ref{alg:sktr} is 
inspired by an alignment-based conformance checking approach~\cite{bogdanov2022conformance}. It takes as input a stochastic trace model $T$ and a process model $SN$, constructs (Line~\ref{algline:SSP}) a stochastic synchronous product $SSP$ (see~\cite{bogdanov2022conformance} for a formal definition) and a corresponding reachability multigraph $RG$ (Line~\ref{algline:RG}, Section~\ref{sec:RG}), and searches for an optimal alignment (Lines~\ref{algline:start_search}--\ref{algline:end_search}, Sections~\ref{sec:search_rg}--\ref{sec:integrating seq probs within SKTR}). The initialization stage involves hyperparameter settings based on the guidelines presented in Section~\ref{sec:model} for selecting the cost function, calculations to determine the sequence--probability mapping, and via exploration--exploitation, testing and grid search (see examples in~\cite{chocron2022delay}) for the history length and trade-off parameters. 

\begin{algorithm}[htpb]
	\SetKwInOut{Input}{input} \SetKwInOut{Output}{output} \SetKwInOut{Initialization}{initialization}
	\SetAlgoLined
	\Input{process model $SN$; stochastically known trace $T$}
	\Output{deterministically known trace}
	\Initialization{set hyperparameter values: cost function $C$, sequence--probability mapping $P$, history length and trade-off parameter $\alpha$.} 
	\BlankLine
	generate stochastic synchronous product $SSP$ from $SN$ and $T$\; \label{algline:SSP}
	generate the reachability multigraph $RG$\; \label{algline:RG}
	\tcp{search ($Dijkstra$, or $A^*$) over $RG$ for shortest (cheapest) path from initial to final markings, $m_i$ to $m_f$, respectively}
	\For {\textbf{each} $e' \in RG$ that is encountered during the search \label{algline:start_search}} { 
		\If{$e'$ corresponds to an $SSP$'s nonsynchronous move \label{algline:nonsynchronous_start}} 
		{
			$Weight(e') \leftarrow 1$\;
		} \label{algline:nonsynchronous_end}
		\Else
		{\If {if any subsequence $s'$ of the path to $e'$ in $RG$ that ends with the activity label corresponding to the edge $e'$ can be mapped by $P$\;}
			{set $s'$ as the longest subsequence recognizable by $P$\;
				$Weight(e') \leftarrow \alpha*C(e') + (1-\alpha)*C(1-P(s'))$\;
			}
			\Else
			{$Weight(e') \leftarrow C(e')$\;}
		}
	} \label{algline:end_search}
	recovered trace $\leftarrow$ $RG$ path with the cheapest sequence of activity labels \label{algline:recovered}
	\caption{The \gls{SKTR}}
	\label{alg:sktr}
\end{algorithm}

\begin{figure}[htpb]
	\begin{center}	
		\includegraphics[ width=0.45\textwidth, height=1.2in]{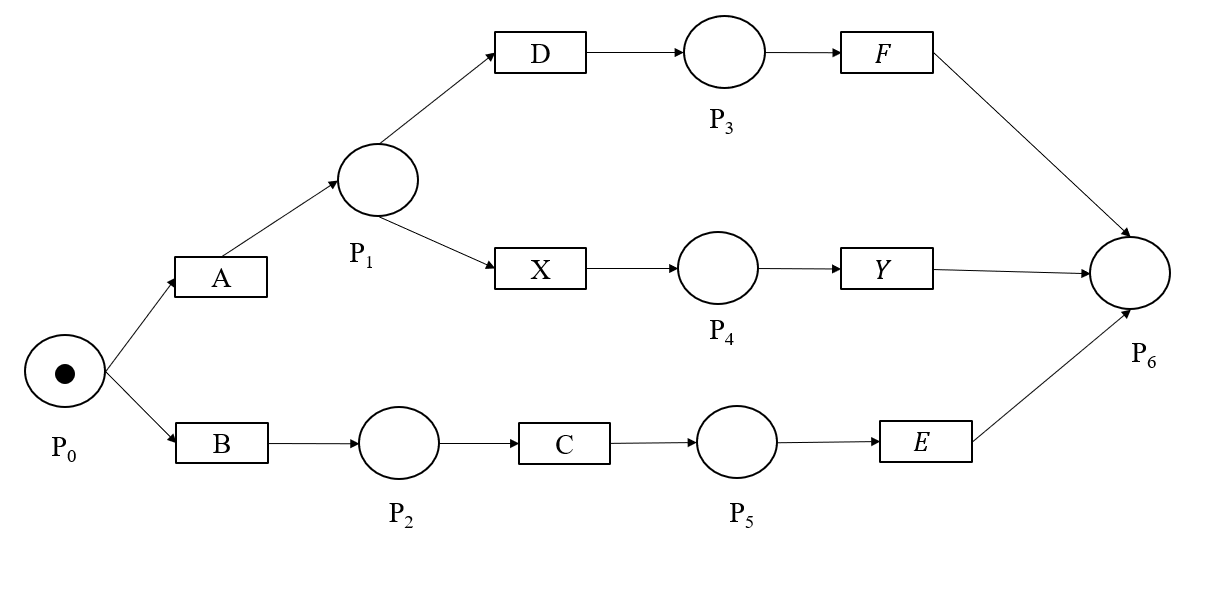}
		\caption{A process model of the running example.}
		\label{fig:process_model_running_example}
	\end{center}
\vskip-.1in
\end{figure}

Consider, for example, the stochastic trace in Table~\ref{tab:stochastic_trace_for_the_running_example}, its stochastic model (Figure~\ref{fig:stochastic log model example}), and the process model in Figure~\ref{fig:process_model_running_example}. \gls{SKTR} recovers %to compute the conformance cost between the trace and the model, and then extracting 
the trace from the optimal alignment by removing the skip symbol aligned with a model's move. The recovered trace $\langle B,C,E\rangle$ is different than the Argmax prediction $\langle A,C,E\rangle$ (see Section~\ref{sec:model}). If the process model %in Figure~\ref{fig:process_model_running_example} 
reflects reality faithfully, we can conclude that the Argmax heuristic offers an inferior prediction since $\langle A,C,E \rangle$ is not valid according to the model. This depends on the process model's quality since a process model with low fitness may not conform with the aforementioned trace, while the trace may be possible in reality. 

\subsection{Constructing a Reachability Multigraph} \label{sec:RG}

We denote the stochastic synchronous product $SSP=(N,m_i,m_f)$, where $N=(P,T,F,\lambda,W)$. Alternative nonsynchronous moves affect the trace recovery accuracy. Thus, we transform $N$, given $m_i$ and $m_f$, into a reachability \textit{multigraph} $RG(N)$ in which optimal alignment is searched over all alternative transitions rather than randomly selecting one that participates in a nonsynchronous move. 
$RG(N)$ is constructed such that two markings $m$ and $m'$, which are two nodes in the graph, have an edge $e'$ between them if and only if there is a corresponding transition $t'$ within the $SSP$. Differently from a deterministic setting,% in which two adjacent places within the trace model are connected via a single edge, 
in an \gls{SK} trace there may be more than a single edge connecting two adjacent places. %leading to a reachability multigraph $RG(N)$. For example, in Figure~\ref{fig:reachability_graph}, which presents the reachability multigraph of the $SSP$ of our running example, the two nonsynchronous moves $(>>,A)$ and $(>>,B)$ start from the initial marking and end in the same adjacent marking. 

\begin{figure}[htpb]
	\begin{center}	
		\includegraphics[ width=0.45\textwidth, height=1.1in]{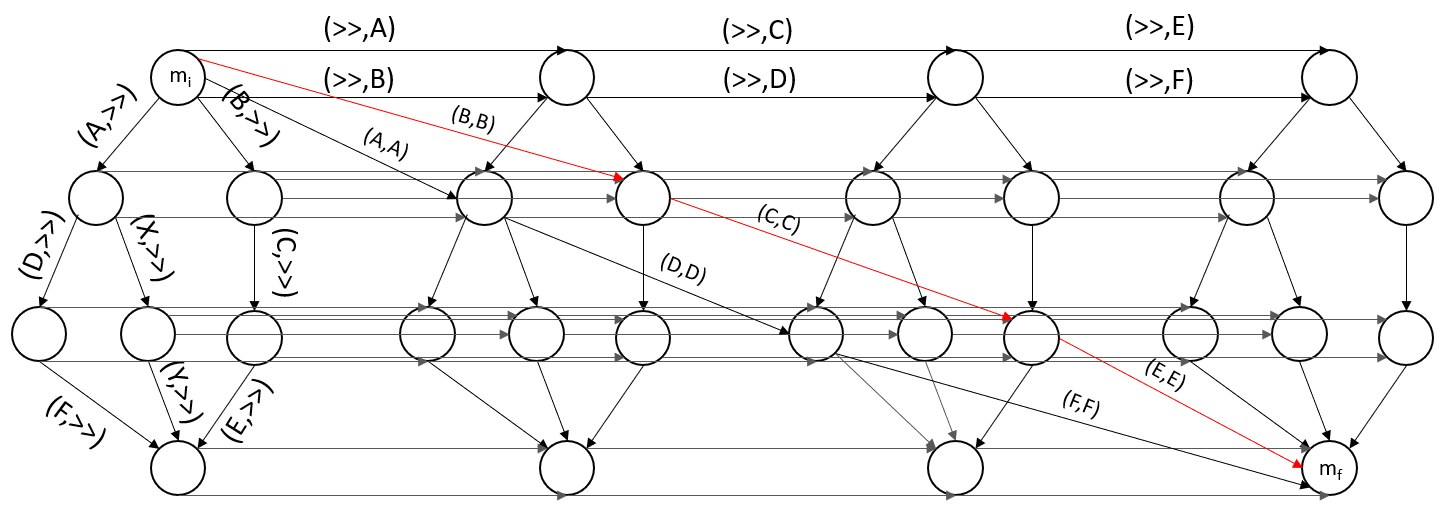}
		\caption{Reachability multigraph for the $SSP$ of the running example. The shortest path is in red.} %(exponential cost function)}
		\label{fig:reachability_graph}
	\end{center}
\vskip -.1in
\end{figure}

\subsection{Searching the Reachability Multigraph} \label{sec:search_rg}
The algorithm searches over the reachability multigrpah $RG$ (Lines~\ref{algline:start_search}-\ref{algline:end_search}) for the shortest (cheapest) path between $m_i$ and $m_f$, using $Dijkstra$ or $A^*$, while enabling transition differentiation by altering edge weights (costs) within the the graph. 
During the search, \algNameNewSpace evaluates the multigraph edges. A weight of $1$ is assigned to edges that correspond to nonsynchronous moves. The edge weights that correspond to synchronous moves are computed according to their probability in the $SSP$ and the chosen cost function. Each time the algorithm encounters such an edge, it computes a final weight based on its original weight and the probability of the sequence of activities that lead to this edge (see Section~\ref{sec:computing seq probs}), which is added to the overall alignment weight for the selected edge. The search procedure, when executed on the running example, yields the path marked in red in Figure~\ref{fig:reachability_graph}. 

\subsection{Computing Sequence--Probability}\label{sec:computing seq probs}
We now show how \gls{SKTR} uses history to improve trace recovery accuracy. Conformance checking implicitly assumes that to compute the cost of a transition it is sufficient to know the preceding marking in the synchronous product and, in a stochastic setting, also the probability associated with the considered trace event. This assumption ignores the likelihood of an activity happening, which depends on activities that have already happened and their ordering.  We should, therefore, take into account the firing sequence of preceding transitions. Take for example a simple process model that permits four firing sequences: $\langle A,B,C \rangle$, $\langle A,B,D \rangle$, $\langle B,A,C \rangle$, and $\langle B,A,D \rangle$. The underlying Petri net of such a process would allow for activities $A$ and $B$ to be executed in any order, after which activity $C$ or $D$ would occur. Based only on the process model, one cannot predict the transition that follows $A$ and $B$ (in any order) better than a random guess. On the other hand, assuming that trace $\langle A,B,C \rangle$ occurred 99 times, $\langle A,B,D \rangle$ once, $\langle B,A,D \rangle$ 98 times, and the $\langle B,A,C \rangle$ appeared twice, it would be much more likely that $C$ would following the sequence $\langle A,B \rangle$, and $D$ would follow $\langle B,A \rangle$. Probabilities of such sequences could be estimated from the process log and used during the search for an optimal alignment, as discussed next.

\begin{sloppypar} 
Let $s = \langle t_1, t_2, .., t_n\rangle$ be a transition firing sequence. Let $H_{i}(t_{j})$ be the set of all observed sub-sequences between $t_{j-i}$ and $t_{j-1}$ followed by $t_j$,  ($\{\langle t_{j-i}, t_{j-i+1},.., t_{j} \rangle, \langle t_{j-i+1}, t_{j-i+2},.., t_{j} \rangle ,.., \langle t_{j-1}, t_{j} \rangle \}$ where $i < j$). By extension, $H_{i}(s)$ is the set of all sub-sequences in $s$ of size up to $i+1$ (the computational effort for each trace is bounded by $|n\cdot (i+1)|$). Given a supervised part of a log, we can compute occurrence probabilities of all sub-sequences, termed {\em histories}. To illustrate, consider a log with two traces, $s_1 = \langle A,A,B,C\rangle$ and $s_2 = \langle A,B,B,C\rangle$. By arbitrarily choosing $i=2$ we get $H_{2}(s_1)=\{ \langle A,A\rangle, \langle A,A,B\rangle, \langle A,B\rangle, \langle A,B,C\rangle, \langle B,C\rangle \}$, and $H_{2}(s_2)=\{ \langle A,B\rangle, \langle A,B,B\rangle, \langle B,B\rangle, \langle B,B,C\rangle, \langle B,C\rangle \}$. Note that the sum of probabilities for each history with identical length and labels for the sequence between $t_{j-i}$ and $t_{j-1}$ is $1$, allowing to calculate the conditional probability for each observed $t_j$ label. The sequence--probability mapping $P$ for our example is:
\begin{align*}
    &P(A|\langle A\rangle)=\tfrac{1}{3}, \quad P(B|\langle A \rangle)=\tfrac{2}{3}; && \\
    &P(C|\langle A,B \rangle)=\tfrac{1}{2},  P(B|\langle A,B \rangle)=\tfrac{1}{2};&&\\
    &P(A|\langle A,B \rangle)=1;\\
    &P(C|\langle B \rangle)=\tfrac{2}{3}, \quad P(B|\langle B \rangle) =\tfrac{1}{3};\\
    &P(C|\langle B,C \rangle)=1.
\end{align*}
The probability of any sub-sequence ({\em e.g.}, $P(\langle t_1, t_2, t_3 \rangle)$) is $\frac{\#\langle t_1, t_2, t_3\rangle}{\#\langle t_1, t_2\rangle}$ where $\#$ refers to the number of times such a sequence appears in the log. 
\end{sloppypar}

\subsection{Using Sequence--Probability Mapping Within \algNameNew}\label{sec:integrating seq probs within SKTR}
To demonstrate how to use probability mapping for trace recovery, consider edge $e'$, which corresponds to activity $C$, and assume that the path to $e'$ was preceded by three edges, corresponding to activities $A,A,B$ in that order. If $e'$ is a nonsynchronous move, its weight would be 1 for any preceding activity sequence (Lines~\ref{algline:nonsynchronous_start}--\ref{algline:nonsynchronous_end}). Otherwise, $e'$ corresponds to a synchronous move -- in which case, we look for the longest available sequence of $e'$'s preceding activities in our sequence--probability mapping. For example, based on the sequence probabilities computed in Section \ref{sec:computing seq probs}, the sub-sequence $\langle A, A, B, C \rangle$ is not recognizable by $P$ since $i=2$. When we remove the farthest away activity that precedes $e'$ from the sequence, we get $\langle A, B, C \rangle$, which is recognized by $P$ with a probability $0.5$. The final weight of the edge is computed as follows: $Weight(e')= \alpha*C(e') + (1-\alpha)*C(1-P(s'))$ where $0 \leq \alpha\leq 1$ is a hyperparameter that allocates the cost of the edge between its probability and the path from which it was reached, and $C$ is the selected cost function. If we do not want to consider history (i.e., $i=0$) or no sub-sequence was recognized by $P$, then the cost of the edge is determined by its probability  $Weight(e')=C(e')$.        

\section{Empirical Evaluation} \label{sec:experimental_study}
We use five publicly available datasets to evaluate the performance of \algNameNewSpace. %when recovering uncertain traces, 
%. We use the first two, 
With BPI 2012 and BPI 2019 we test our approach by inserting controlled uncertainty similar to earlier works (e.g.,~\cite{bogdanov2022conformance} and~\cite{pegoraro2020efficient}). Three food preparation process videos, namely Breakfast, 50 salads, and GTEA, serve as popular benchmarks for machine learning algorithms. Log uncertainty for these benchmarks stems from applying a state-of-the-art machine learning algorithm to segment a temporally untrimmed video by time and label each segment with an activity. To the best of our knowledge, this paper is the first to apply process mining to real-life uncertain datasets. 

We adopt the widely accepted accuracy performance measure as the main performance measure for this task. % since many prediction models and sensors are not perfectly accurate. This leads to true labels that are not necessarily associated with the highest probabilities and thus the Argmax heuristic may lead to inaccurate trace recovery.
%\ag{add here a paragraph on measures of evaluation}
Overall performance comparison (Section~\ref{SKTR_Performance}) is followed by a separate discussion of experiments on the BPI datasets (Section~\ref{sec:BPI_datasets}) and the video datasets (Section~\ref{sec:Video_datasets}). 

\subsection{\algNameNewSpace Overall Performance} \label{SKTR_Performance}
Table~\ref{tab:table2} compares the performance of \gls{SKTR} (Algorithm~\ref{alg:sktr}) and the Argmax heuristic (with ASFormer network~\cite{yi2021asformer} for the video datasets) over 30 experiments. \gls{SKTR} shows a significant average accuracy improvement over the baseline method (p-values of paired one-sided t-tests were $\ll 0.05$) with an improvement of about $10\%$ on average across all datasets.

%\begin{scriptsize}
\begin{table}[htbp]
	\centering
%        \begin{adjustbox}{width=0.5\textwidth}
		\begin{tabular}{l c c c c c} 
		\hline
					\tiny{\textbf{Algorithm}} & \tiny{Breakfast} & \tiny{50Salads} & \tiny{GTEA}  & \tiny{BPI2012} & \tiny{BPI2019}\\ 
		\hline
		\tiny{Argmax} & \tiny{0.70} & \tiny{0.83} & \tiny{0.73} & \tiny{0.78} & \tiny{0.80}  \\ % <--
		\hline
		\tiny{\algNameNewSpace} & \tiny{\textbf{0.81}} & \tiny{\textbf{0.89}}  & \tiny{\textbf{0.79}} & \tiny{\textbf{0.92}} & \tiny{\textbf{0.82}} \\ 
		\hline
		\tiny{improvement} & \tiny{15.7\%} & \tiny{7.2\%} & \tiny{8.2\%} & \tiny{17.9\%} & \tiny{2.5\%}  \\% <--
		\hline
		\vspace{5pt}
	\end{tabular}
	\vskip-.1in
%        \end{adjustbox}
	\caption{\label{tab:table2} Trace recovery accuracy and improvement across five datasets using the Argmax heuristic and \algNameNew.}
	\vskip-.1in
\end{table}
%\end{scriptsize}

%\vspace{-10pt}
\subsection{Experiments with BPI Datasets} \label{sec:BPI_datasets}
Experiments with the BPI datasets were designed to reveal insights about \algNameNewSpace performance under varying levels of uncertainty, model quality, and different cost functions. %may perform similarly or significantly change the performance of the suggested trace recovery approach. 
Due to space considerations, we present results for the BPI 2012 dataset. Data preparation procedures and results were similar for BPI 2019.

\subsubsection{Data Preparation and Preprocessing} \label{sec:BPI_preprocess}
To prepare the uncertain log, we controlled the following parameters:
\begin{compactenum}[(1)]
    \item $P_a$, which sets the probability of the original transition to be assigned the highest probability among its set of alternative transitions. For example, setting $P_a=0.5$ will result in each original transition being assigned the highest probability approximately half of the time; otherwise, the highest probability is assigned to an alternative transition.
    \item $T_s$ is the number of traces used for process model discovery. Typically, low $T_s$ values would lead to low fitness and high precision, and high values would increase fitness and decrease precision. 
\end{compactenum}
We randomly chose $T_s$ traces for process model discovery, and another 2000 random traces to test \gls{SKTR} (and Argmax). We generated noise within each trace by adding to each transition within it all activity labels as alternative transitions. The probability mass was divided among the set of alternative transitions according to $P_a$ starting with $P_a=0$. In each iteration we added a probability of 0.05 to $P_a$ and regenerated the noise within each trace. %using the new $P_a$ value.  
In essence, we start from a \gls{DK} trace and add noise in a structured way. Then, we evaluate whether the \gls{SK} traces can be recovered by applying \gls{SKTR}. We compared the results with the Argmax heuristic that chooses from each set of alternative transitions the one with the highest probability. Argmax is commonly used to resolve the probability matrix in deep neural networks and to determine their final classification. We evaluate the recovery performance based on the accuracy measure as follows -- given an \gls{SK} trace, we determine its recovered trace using \gls{SKTR} or Argmax and compare each recovered trace to its original \gls{DK} trace. Accuracy is calculated by averaging the recovery accuracy of each trace. We repeated each setting 30 times and averaged the results.

\subsubsection{Results} \label{sec:BPI_results}
\begin{figure*}[htpb]
	\centering
	\begin{subfigure}[b]{0.32\textwidth}
		\centering
		\includegraphics[width=\textwidth, height=3cm]{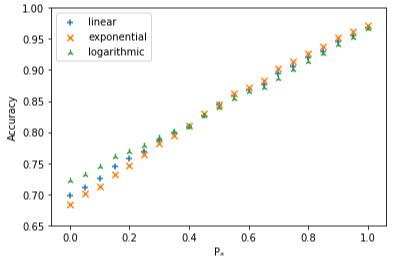}
		\caption{}
		\label{fig:comparing_cost_function_by_pa}
	\end{subfigure} 
	\hfill   
	\begin{subfigure}[b]{0.32\textwidth}
		\centering
		\includegraphics[width=\textwidth, height=3cm]{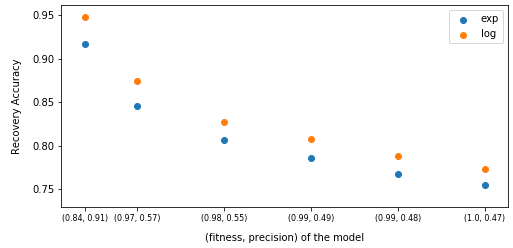}
		\caption{}
		\label{fig:model_discovery_pa_0.3}
	\end{subfigure} 
	\begin{subfigure}[b]{0.32\textwidth}
		\centering
		\includegraphics[width=\textwidth, height=3cm]{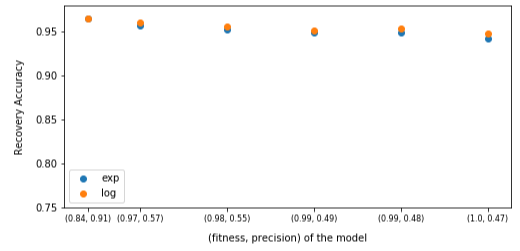}
		\caption{}
		\label{fig:model_discovery_pa_0.8}
	\end{subfigure} 
	\caption{Recovery accuracy by \algNameNewSpace (a) as a function of $P_a$. %for different cost functions
 (b) and (c) present logarithmic and exponential cost functions for low ($P_a=0.1$) and reasonable ($P_a=0.8$) log qualities, respectively, as a function of the number of traces that were used for model discovery (presented as model's fitness and precision). \label{fig:bpi_experiments}}
	%\vskip-.1in
\end{figure*}

Figure~\ref{fig:bpi_experiments} summarizes the results and identifies conditions under which a specific cost function may affect trace recovery performance. 
Figure \ref{fig:comparing_cost_function_by_pa} compares cost functions under varying log qualities (varying $P_a$ values) and a model that was discovered by dusing $T_s=15$ (preliminary experiments indicate that this number achieves high recovery accuracy). A logarithmic cost function performed favorably compared to an exponential function for an unreliable log (low $P_a$ values), across a wide range of models (see, Figure \ref{fig:model_discovery_pa_0.3}). Performance deteriorated with a decrease in the model's precision. For high quality logs manifested by high $P_a$ values, the selection of a cost function did not make a significant difference with a slight deterioration in recovery accuracy as the model precision decreases (see Figure \ref{fig:model_discovery_pa_0.8}). % we find that the cost function affected This pattern was not preserved for  (i.e., a high quality log) in which case model's precision did not such an effect. In this case, models with higher fitness but lower precision had an unfavourable effect when combined with the exponential function. The opposite was true for the logarithmic cost function (see Figure \ref{fig:model_discovery_pa_0.8}). The exact interplay between the quality of the model and the log, and its effect on the recovery accuracy of \algNameNew is still left to be studied in future research. 

Performance differences can be traced back to the cost function properties. Exponential cost function inflict higher penalties on lower probabilities compared to logarithmic functions (see, Figure \ref{fig:cost_func_trustworth}). Therefore, a logarithmic function may favor alignments with as many synchronous moves as possible (even if their probabilities are low), whereas an exponential function may prefer alignments with nonsynchronous moves over ones with multiple low probability synchronous moves.  
Consider, for example, a model in which two alignments are being evaluated by \algNameNew. Alignment 1 consists of five synchronous moves and a single nonsynchronous move, while Alignment 2 consists of four synchronous moves and two nonsynchronous moves. If the probabilities associated with the synchronous moves of Alignment 1 are lower than those of Alignment 2, \algNameNew with a logarithmic function is more likely to choose Alignment 1 since low probabilities (though not extremely low) would not induce a high cost for the synchronous moves, compared to Alignment 2 for which an additional nonsynchronous move would induce a cost of 1. In contrast, \algNameNew~with an exponential function may prefer Alignment 2 as the cost induced by five low probability synchronous moves can exceed the cost induced by the four synchronous moves of Alignment 2 even after considering the additional nonsynchronous move in Alignment 2. Choosing highly probable transitions, even with the additional cost of nonsynchronous moves, can lead to recovery errors if these transitions are not the true ones. Thus, for logs with a low quality and a reasonable model, one would wish to trust the process model while minimizing nonsynchronous moves. 

As $P_a$ values increase, different cost functions exhibit similar performance since the functions that assign higher penalties would choose the original transitions more often given that these transitions currently have higher probabilities.
Based on the above explanations, Figures \ref{fig:model_discovery_pa_0.3} and \ref{fig:model_discovery_pa_0.8} present the results of experiments with a varying model quality (by changing $T_s$) and low ($P_a=0.1$) and high ($P_a=0.8$) log quality, respectively. When the log is prone to errors (Figure~\ref{fig:model_discovery_pa_0.3}), a logarithmic cost function led to slightly better performance of \algNameNew~compared to an exponential cost function. This happens because a logarithmic function prioritizes alignments with fewer nonsynchronous moves and relies more on the model compared to the stochastic one. The x-axis shows that as $T_s$ increases and model's precision decreases, \algNameNewSpace performance decreases. As seen in Figure~\ref{fig:model_discovery_pa_0.8}, for a reasonable log quality ($P_a=0.8$) cost functions perform almost equally. %presents a case with with reasonable log quality , thus the model is less important, and both functions often produce similar alignments since the true activity labels in the log are likely to have the highest probability, favoring their selection by \algNameNew~(this can also be observed in Figure~\ref{fig:comparing_cost_function_by_pa} for high values of $P_a$). Choosing the right cost function may improve trace recovery accuracy by several percentages, especially for unreliable logs. 

\subsection{Experiments with Video Datasets} \label{sec:Video_datasets}

We used three publicly available popular machine learning datasets, % \href{https://serre-lab.clps.brown.edu/resource/breakfast-actions-dataset/}{Breakfast}, \href{https://cvip.computing.dundee.ac.uk/datasets/foodpreparation/50salads/}{50 Salads}, and \href{https://cbs.ic.gatech.edu/fpv/}{GTEA}, 
which contain videos of food preparation. These benchmarks are used to compare activity segmentation algorithms, predicting the correct activity label for timestamps within each video stream.

\subsubsection{Data Preparation and Preprocessing} \label{sec:Video_preprocess}

To retrieve labels' probabilities from videos, we reproduced the results of a state-of-the-art deep learning network for action segmentation, ASFormer \cite{yi2021asformer}, which receives as input a video stream and assigns an activity label for each selected timestamp. The accuracy of the model ranges in 70\%--86\% depending on the dataset and selected parameters. For a fair comparison, we used the default network parameters. Prior to the assignment of labels, for each timestamp, the model generates the probability of each label being the correct one in the form of a matrix (termed {\em softmax matrix}) where rows correspond to the possible labels and columns to timestamps.  Then, it assigns to each timestamp the label with the highest probability (Argmax). %We refer to the probability matrices as the 
The softmax matrices are used them as inputs to \algNameNew. %In other words, we replace the deep network final layer (Argmax) by \algNameNewSpace for the same input. 
In addition, \algNameNewSpace receives a training set of several labeled videos from which it discovers a reference process model. \algNameNewSpace converts the softmax matrices into \gls{SK} traces, constructs a reachability multigraph between the process model and each matrix, and recovers the labels using Algorithm \ref{alg:sktr}.

After applying ASFormer %on each dataset 
and extracting a softmax matrix for each video, we sampled about 30-60 timestamps from each video %to reduce the computational burden of aligning long traces) 
and passed the samples to \algNameNew. We compared \algNameNew's output against the ground truth labels, annotated by humans %(i.e., the true \gls{DK} traces) 
and computed its accuracy. %, which as mentioned, is a common metric of evaluation on these benchmarks.  

\subsubsection{Results}\label{sec:video_results} 
The results, summarized in Table~\ref{tab:table2}, indicate that \algNameNewSpace significantly outperforms the baseline deep learning network with a relative accuracy improvement of $7.2\%$-$15.7\%$. %($2.5\%$-$17.9\%$ for the BPI datasets). 
Using the sequence--probability feature that considers the path (history) containing previous activities (Section \ref{sec:integrating seq probs within SKTR}) improved the accuracy by $\sim 1\%$-$2\%$ compared to the standard practice of considering only the current marking. We believe that developing integration procedures of history with \algNameNewSpace can improve the results even more but this is outside the scope of this research. The biggest advantage \algNameNewSpace offers in comparison to the baseline %(i.e., Argmax) 
is its reference process model that is utilized during the label recovery process. The algorithm considers the process perspective, enabling it to predict the correct labels by %not only observing each individual label and its corresponding probability, but also by 
considering sequences of labels and how well they are aligned with the process. 

\section{Related Work} \label{sec:related_work}
%The notion of \gls{SK} event logs was implied by
\citeauthor{suciu2011probabilistic}~\cite{suciu2011probabilistic}  describe probabilistic databases as ``...databases where the value of some attributes or the presence of some records are uncertain and known only with some probability.'' Commonly, data uncertainty is classified into attribute- or record-level uncertainty~\cite{suciu2011probabilistic, aggarwal2008survey}, where the former term refers to an uncertain event attribute for which the possible values are known with probability, and the latter to the probability that a record (a trace) exists. 
In the context of this work we focus on attribute-level uncertainty. 

A rich body of research focuses on (semi-)automatically identifying data quality issues, %cleaning and correcting them, 
{\em e.g.},~\cite{suriadi2017event,wang2015cleaning}. These studies focus on identifying atypical data patterns with respect to other data or a process and developing repair mechanisms but do not exploit probabilistic knowledge %about the uncertainty 
of data values. %to solve data quality issues. 
%Rather, the . 
Some studies, to improve model discovery results, filter out an event log. For example, \citeauthor{conforti2016filtering}~\cite{conforti2016filtering} develop a threshold-based framework for filtering out infrequent behaviors. Mixed-integer linear programming is used to find an automaton with a minimal number of arcs that represents the log behavior, which is used for finding frequency thresholds. The events that do not fit the automaton, subject to the threshold, are filtered out. \citeauthor{rogge2013repairing}~\cite{rogge2013repairing} present an alignment-based technique to fill in missing trace events. Among a set of optimal alignments, the one with the highest probability is selected. In contrast, we factor the probabilities of different paths during the search and return a single optimal alignment rather than a set of alignments. Another difference, is that we compute alignments with respect to an \gls{SK} trace.
   
\citeauthor{sani2017improving}~\cite{sani2017improving} propose techniques to detect incomplete traces or traces that do not terminate properly by identifying the probability in which an activity appears in the context of a specific behavior. For example, how likely it is that activity $d$ appears after the sequence $\langle a,b,c \rangle $. The conditional probabilities can be then filtered according to a threshold.
\citeauthor{van2020partial}~\cite{van2020partial} focus on the ordering of events when some of their timestamps are imprecise due to missing, non-synchronized, or manual recordings of date/time information (also called order uncertainty). The authors test three different models in which the probability of a trace variation or a specific sequence of events are estimated based on event-log traces that do exhibit order uncertainty (deterministic traces). 
A recent paper by \citeauthor{andrews2020quality} \cite{andrews2020quality}, who take a universal approach to identifying data by developing a semi-automated data quality evaluation framework of event data extracted from relational databases. Their approach is based on the development of quantitative indices for measuring 12 different data quality dimensions (e.g., precision, completeness, uniqueness). 
 
\citeauthor{pegoraro2019mining}~\cite{pegoraro2019mining} classify data uncertainty into one of two classes: strong and weak. The former states that an event's attribute may have several possible values but their probabilities of being the true value are unknown. The latter assumes that the probability distribution over possible values is known.  \citeauthor{pegoraro2019discovering} \cite{pegoraro2019discovering} develop a process discovery technique for logs with strong uncertainty. The technique constructs a directly-follows graph, extracts a process model, and applies filters to simplify it. \citeauthor{pegoraro2020efficientspaceandtime}~\cite{pegoraro2020efficientspaceandtime} construct a behavior graph, which accommodates precedence relationships among events for logs with strong uncertainty, allowing the discovery of models from logs using directly-follows relationships. \citeauthor{pegoraro2021conformance} \cite{pegoraro2021conformance} evaluate lower and upper bounds on the cost of alignment-based conformance under the strong uncertainty assumption. \citeauthor{bergami2021tool} \cite{bergami2021tool} develop a conformance checking technique between a stochastic workflow network and a deterministic trace. Their technique filters model paths having occurrence probabilities higher than a predetermined threshold and ranks their conformance score with respect to the reference trace as a product of a distance measure (e.g., Levenshtein distance) and path's probability.
%\citeauthor{cohen2021uncertain} characterize different uncertainty settings for a process model and an event log~\cite{cohen2021uncertain}. The authors refer to traces in which attribute values are either certain (denoted as \gls{DK} traces) or \gls{SK} with a probability distribution function that describes each set of attribute's possible values and their respective probabilities.
\citeauthor{bogdanov2022conformance} \cite{bogdanov2022conformance} propose an alignment-based algorithm that computes the conformance cost between a model and an \gls{SK} trace. 
%This paper proposes \gls{SKTR}, a trace recovery algorithm from \gls{SK} traces. 

Differently from previously introduced solutions~\cite{bogdanov2022conformance}, in \gls{SKTR} the level of trust in model and log is adjusted according to their qualities via selection of different cost functions. Moreover, recovery of activity labels can consider the probabilities of activity sequences. To restore activity labels, the synchronous product is modeled as a \textit{multigraph}, which enables distinguishing between nonsynchronous moves, a novel feature of the algorithm. 

\section{Conclusions and Future Work} \label{sec:conclusions}
We propose accurate trace recovery process from uncertain data, a crucial task for credible process analysis and decision making. %with the increasing number of sensors and data generated by predictive models. 
We focus on \gls{SK} event logs in which activity labels are only known in stochastic terms, a setting characteristic of neural networks and sensor outputs. The suggested approach calculates the conformance between a process model and an \gls{SK} trace and recovers, as the true trace, the best alignment within this stochastic trace. Results of experiments over five datasets, three of which document processes via videos, indicate significant trace recovery accuracy improvements of $2.5-17.9\%$ by \gls{SKTR} compared to the common Argmax heuristic and a deep learning network. 
For future work, we plan to explore computationally efficient techniques for computing alignments with long \gls{SK} traces from long videos/processes. Another research direction involves exploring additional types of cost functions and integrating history into \algNameNew.  

\printbibliography

@article{aggarwal2008survey,
  title={A survey of uncertain data algorithms and applications},
  author={Aggarwal, Charu C and Philip, S Yu},
  journal={IEEE Transactions on Knowledge and Data Engineering},
  volume={21},
  number={5},
  pages={609--623},
  year={2008},
  publisher={IEEE}
}

@article{andrews2020quality,
  title={Quality-informed semi-automated event log generation for process mining},
  author={Andrews, Robert and van Dun, Christopher GJ and Wynn, Moe Thandar and Kratsch, Wolfgang and R{\"o}glinger, MKE and ter Hofstede, Arthur HM},
  journal={Decision Support Systems},
  volume={132},
  pages={113265},
  year={2020},
  publisher={Elsevier}
}

@article{beerepoot2023biggest,
  title={The biggest business process management problems to solve before we die},
  author={Beerepoot, Iris and Di Ciccio, Claudio and Reijers, Hajo A and Rinderle-Ma, Stefanie and Bandara, Wasana and Burattin, Andrea and Calvanese, Diego and Chen, Tianwa and Cohen, Izack and Depaire, Beno{\^\i}t and others},
  journal={Computers in Industry},
  volume={146},
  pages={103837},
  year={2023},
  publisher={Elsevier}
}

@inproceedings{bogdanov2022conformance,
  title={Conformance Checking Over Stochastically Known Logs},
  author={Bogdanov, Eli and Cohen, Izack and Gal, Avigdor},
  booktitle={Business Process Management Forum: BPM 2022 Forum, M{\"u}nster, Germany, September 11--16, 2022, Proceedings},
  pages={105--119},
  year={2022},
  organization={Springer}
}

@article{cohen2021uncertain,
  title={Uncertain Process Data with Probabilistic Knowledge: Problem Characterization and Challenges},
  author={Cohen, Izack and Gal, Avigdor},
  journal={Proceedings of the International Workshop Problems21, co-located with the 19th International Conference on Business Process Management BPM 2021, Italy, published in CEUR Workshop Proceedings},
  volume={2938},
  pages={51--56},
  year={2021}
}

@article{chocron2022delay,
  title={Delay Prediction for Managing Multiclass Service Systems: An Investigation of Queueing Theory and Machine Learning Approaches},
  author={Chocron, Elisheva and Cohen, Izack and Feigin, Paul},
  journal={IEEE Transactions on Engineering Management},
  year={2022},
  publisher={IEEE}
}

@article{conforti2016filtering,
  title={Filtering out infrequent behavior from business process event logs},
  author={Conforti, Raffaele and La Rosa, Marcello and ter Hofstede, Arthur HM},
  journal={IEEE Transactions on Knowledge and Data Engineering},
  volume={29},
  number={2},
  pages={300--314},
  year={2016},
  publisher={IEEE}
}

@inproceedings{bergami2021tool,
  title={A Tool for Computing Probabilistic Trace Alignments},
  author={Bergami, Giacomo and Maggi, Fabrizio Maria and Montali, Marco and Pe{\~n}aloza, Rafael},
  booktitle={International Conference on Advanced Information Systems Engineering},
  pages={118--126},
  year={2021},
  organization={Springer}
}

@inproceedings{pegoraro2019mining,
  title={Mining uncertain event data in process mining},
  author={Pegoraro, Marco and van der Aalst, Wil MP},
  booktitle={2019 International Conference on Process Mining (ICPM)},
  pages={89--96},
  year={2019},
  organization={IEEE}
}

@article{pegoraro2021conformance,
  title={Conformance checking over uncertain event data},
  author={Pegoraro, Marco and Uysal, Merih Seran and van der Aalst, Wil MP},
  journal={Information Systems},
  volume={102},
  pages={101810},
  year={2021},
  publisher={Elsevier}
}

@inproceedings{pegoraro2019discovering,
  title={Discovering process models from uncertain event data},
  author={Pegoraro, Marco and Uysal, Merih Seran and Van Der Aalst, Wil},
  booktitle={International Conference on Business Process Management},
  pages={238--249},
  year={2019},
  organization={Springer}
}

@inproceedings{pegoraro2020efficient,
  title={Efficient construction of behavior graphs for uncertain event data},
  author={Pegoraro, Marco and Uysal, Merih Seran and Van Der Aalst, Wil},
  booktitle={International Conference on Business Information Systems},
  pages={76--88},
  year={2020},
  organization={Springer}
}

@article{pegoraro2020efficientspaceandtime,
  title={Efficient Time and Space Representation of Uncertain Event Data},
  author={Pegoraro, Marco and Uysal, Merih Seran and Van Der Aalst, Wil},
  journal={Algorithms},
  volume={13},
  number={11},
  pages={285},
  year={2020},
  publisher={Multidisciplinary Digital Publishing Institute}
}

@book{rogge2013repairing,
  title={Repairing event logs using stochastic process models},
  author={Rogge-Solti, Andreas and Mans, Ronny S and van der Aalst, Wil MP and Weske, Mathias},
  volume={78},
  year={2013},
  publisher={Universit{\"a}tsverlag Potsdam}
}

@inproceedings{sani2017improving,
  title={Improving process discovery results by filtering outliers using conditional behavioural probabilities},
  author={Sani, Mohammadreza Fani and van Zelst, Sebastiaan J and Van Der Aalst, Wil},
  booktitle={International Conference on Business Process Management},
  pages={216--229},
  year={2017},
  organization={Springer}
}

@article{suriadi2017event,
  title={Event log imperfection patterns for process mining: Towards a systematic approach to cleaning event logs},
  author={Suriadi and Andrews, Robert and Ter Hofstede, Arthur HM and Wynn, Moe Thandar},
  journal={Information Systems},
  volume={64},
  pages={132--150},
  year={2017},
  publisher={Elsevier}
}

@article{suciu2011probabilistic,
  title={Probabilistic databases, synthesis lectures on data management},
  author={Suciu, Dan and Olteanu, Dan and R{\'e}, Christopher and Koch, Christoph},
  journal={Morgan \& Claypool},
  year={2011}
}

@article{van2020partial,
  title={Partial order resolution of event logs for process conformance checking},
  author={Van der Aa, Han and Leopold, Henrik and Weidlich, Matthias},
  journal={Decision Support Systems},
  volume={136},
  pages={113347},
  year={2020},
  publisher={Elsevier}
}

@inproceedings{wang2015cleaning,
  title={Cleaning structured event logs: A graph repair approach},
  author={Wang, Jianmin and Song, Shaoxu and Lin, Xuemin and Zhu, Xiaochen and Pei, Jian},
  booktitle={2015 IEEE 31st International Conference on Data Engineering},
  pages={30--41},
  year={2015},
  organization={IEEE}
}

@article{yi2021asformer,
  title={Asformer: Transformer for action segmentation},
  author={Yi, Fangqiu and Wen, Hongyu and Jiang, Tingting},
  journal={arXiv preprint arXiv:2110.08568},
  year={2021}
}

@InProceedings{10.1007/978-3-319-45348-4_11,
author="Rogge-Solti, Andreas
and Senderovich, Arik
and Weidlich, Matthias
and Mendling, Jan
and Gal, Avigdor",
editor="La Rosa, Marcello
and Loos, Peter
and Pastor, Oscar",
title="In Log and Model We Trust? A Generalized Conformance Checking Framework",
booktitle="Business Process Management",
year="2016",
publisher="Springer International Publishing",
address="Cham",
pages="179--196",
isbn="978-3-319-45348-4"
}

@ARTICLE{9234741,
  author={Janiesch, Christian and Koschmider, Agnes and Mecella, Massimo and Weber, Barbara and Burattin, Andrea and Di Ciccio, Claudio and Fortino, Giancarlo and Gal, Avigdor and Kannengiesser, Udo and Leotta, Francesco and Mannhardt, Felix and Marrella, Andrea and Mendling, Jan and Oberweis, Andreas and Reichert, Manfred and Rinderle-Ma, Stefanie and Serral, Estefanía and Song, WenZhan and Su, Jianwen and Torres, Victoria and Weidlich, Matthias and Weske, Mathias and Zhang, Liang},
  journal={IEEE Systems, Man, and Cybernetics Magazine}, 
  title={The Internet of Things Meets Business Process Management: A Manifesto}, 
  year={2020},
  volume={6},
  number={4},
  pages={34-44},
  doi={10.1109/MSMC.2020.3003135}}
\end{document}